\documentclass[10pt, a4paper]{article}
\usepackage{lrec}
\usepackage{graphicx}
\usepackage{tabularx}
\usepackage{soul}

\usepackage{epstopdf}
\usepackage[latin1]{inputenc}

\usepackage{hyperref}
\usepackage{xstring}

\usepackage{graphicx}
\usepackage[table]{xcolor}
\usepackage{comment}
\usepackage{todonotes}
\usepackage{hhline}

\title{Is it worth it? \\ Budget-related evaluation metrics for model selection} 

\name{Filip Klubi\v{c}ka, Giancarlo D. Salton, John D. Kelleher}

\address{School of Computing \\
		Dublin Institute of Technology\\
        \{filip.klubicka, giancarlo.salton\}@mydit.ie\\
        john.d.kelleher@dit.ie\\}

\abstract{
Projects that set out to create a linguistic resource often do so by using a machine learning model that pre-annotates or filters the content that goes through to a human annotator, before going into the final version of the resource. However, available budgets are often limited, and the amount of data that is available exceeds the amount of annotation that can be done. Thus, in order to optimize the benefit from the invested human work, we argue that the decision on which predictive model one should employ depends not only on generalized evaluation metrics, such as accuracy and F-score, but also on the gain metric. The rationale is that, the model with the highest F-score may not necessarily have the best separation and sequencing of predicted classes, thus leading to the investment of more time and/or money on annotating false positives, yielding zero improvement of the linguistic resource. We exemplify our point with a case study, using real data from a task of building a verb-noun idiom dictionary. We show that in our scenario, given the choice of three systems with varying F-scores, the system with the highest F-score does not yield the highest profits. In other words, we show that the cost-benefit trade off can be more favorable if a system with a lower F-score is employed.
\\ \newline  }

\begin{document}

\maketitleabstract

\section{Introduction}\label{s:intro}

Building linguistic resources, such as corpora or dictionaries, can be very labor-intensive, requiring great amounts of work-hours and expert annotation. However, as pointed out by \newcite{ringger-08}, fixed budgets constrain the amount of annotation that can go into the construction of linguistic resources. In many cases, the amount of available data far exceeds the time and money that is available for annotation, so one can only afford to label a subset of the data. Furthermore, for some linguistic resources only a particular (and sometimes rare) subset of the available data is relevant. Indeed, when the target linguistic phenomenon is relatively rare, the amount of time wasted filtering non-relevant data can be considerable. For example, consider the scenario of creating an idiom dictionary: in this context, only idiomatic phrases should be included in the dictionary and any time or effort expended on filtering non-idiomatic phrases is, essentially, an unnecessary cost to the project. 

Given that annotators are paid either by the hour or, more often, by the number of annotations they produce (measured in words, sentences, phrases etc.), researchers are strongly motivated to reduce the annotation time, and thereby a project's annotation costs, without sacrificing the quality and coverage of the resources they are creating. To this end, the annotation process is often supplemented with predictive models that to an extent streamline, speed up and ultimately cheapen the cost of human intervention. In the scenario of creating a linguistic resource where only a subset of the potential data is relevant for inclusion (for example, a dictionary of idioms) these predictive models can be used to pre-filter the examples presented to an annotator, with the goal of maximizing the percentage of positive instances (e.g. idioms) that the annotator reviews, and minimizing the number of false positives (e.g. non-idioms) that the annotator must manually filter and discard. 

In most projects focused on creating linguistic resources there exists a variety of different predictive (machine learning) models that could be used to pre-filter examples before human-annotation. Consequently, the question arises regarding how to choose the best model to aid in the annotation process. Traditionally, predictive models are evaluated based on established evaluation measures that reflect the accuracy of the model, using metrics such as F-score. Such measures are designed for evaluating predicting processes that include instance scoring and thresholding. An example of this would be deciding on the cutoff score between class labels in a binary prediction task (e.g. idiomatic and non-idiomatic phrases).
However, in this paper we argue that, in contexts where annotation budgets are an important consideration and, hence, we can only afford to annotate a subset of our data, the sequence of the data becomes much more important than the application of the threshold. Thus, in such a constrained setting, selecting a pre-filtering prediction model based on a measure of accuracy which evaluates the interplay of the scoring and thresholding process on an entire test set may not be the optimal approach. Instead, we propose that a more suitable metric for model evaluation in this context is \textit{gain}, and the associated measure of \textit{cumulative gain}.

Unlike standard evaluation metrics that evaluate a prediction model's performance on how successfully it scores the examples and applies a threshold, the \textit{gain} and \textit{cumulative gain} measures provide insight into the performance of a model across different subsets of a test set. Furthermore, they are much more suitable for evaluating how the instances in the test set are sequenced.
Hence, these measures provide insight that is useful for a number of different annotation scenarios. In particular there are three scenarios where we argue that \textit{gain} and \textit{cumulative gain} can be invaluable for the decision-making process:
\begin{enumerate}
	\item We can have a fixed budget and can only afford to annotate a set amount of the data. Here we wish to decide which pre-filtering model is the best one to use.
    \item We have a flexible budget and we can afford to annotate a variable amount of the relevant data, but we still wish to spend as little money as possible with comparable results, and we wish to decide which pre-filtering model is the best one to use.
    \item We have already performed some annotation and we want to know if we should stop, or if it is worthwhile to annotate more, given the likely performance of the pre-filtering model over the next segment of data.
\end{enumerate}

\subsection{Related work}
\label{sec:related}

A number of researchers have already examined questions related to the costs associated with annotation of linguistic resources. \newcite{ringger-08} have done work on estimating the cost of corpus annotation as a step towards selecting which annotation environments are most appropriate for a given project. They perform an analysis of annotation costs on the task of correcting part of speech tags in an automatically annotated corpus. Based on these findings, they present a linear model for estimating the hourly cost of annotation for annotators of various skill levels, as well as a model for two granularities of annotation (sentence at a time and word at a time), thus providing informative guidelines for choosing an optimal annotation environment.

Similarly, \newcite{balamurali-12} present an economic model to asses the benefit accruing from the increase in project cost by performing annotation. They examine the relationship between the additional investment in annotation of WordNet senses and the subsequent increase in accuracy scores on the task of sentiment analysis. Instead of evaluating the predictive models conventionally, by comparing their accuracies, they compare expected profits, which are set up in terms of costs and expected returns. They make a comparison of approaches from different economic perspectives - namely which approach yields maximum expected profit, and which approach yields this profit the earliest (meaning less money can be spent overall).
Their focus is to answer the question ``Is the [subsequent] improvement in accuracy significant enough to justify the [additional] cost of annotation?'', or in other words, they wonder ``Should the extra cost of annotation be incurred for the task at all?'' \cite[p. 3090]{balamurali-12}.

Their questions are similar to ours, but we examine a different setting and propose the \textit{gain} metric as the answer to those questions. The remainder of our paper is thus organized as follows: in Section \ref{s:eval} we discuss standard machine learning model evaluation methods and provide a general motivation for the use of the \textit{gain} measure for evaluating pre-filtering models in contexts where budgetary concerns are relevant; in Section \ref{ss:gain} we introduce the \textit{gain} measure; in Section \ref{s:casestudy} we illustrate the benefits of \textit{gain} in the context of an annotation project using a case study (based on actual research data) of building an idiom dictionary; and finally we round up the paper with a conclusion.

\section{Drawbacks with Traditional Model Evaluation Metrics for Annotation Pre-filtering Models}\label{s:eval}

Within a machine learning context, the standard method for evaluating a predictive model is to first split a dataset into a \textit{training set} and a \textit{test set}. The model is induced by applying a machine learning algorithm to the training data. Once the model has been created it is then run on the test set and a measure of the performance of the model is calculated as a function of how often the predictions made by the model for the instances in the test set match the gold-standard labels for these instances. 

There are a variety of different metrics that can be used to calculate the accuracy of a model on a test set. The simplest is metric is simply the raw accuracy of the model, calculated by dividing the count of test instances the model got correct by the total number of instances in the test set. Other measures of accuracy are designed to handle specific requirements or characteristics of a domain. For example, if a prediction model is being trained to discriminate between two outcomes (e.g. spam vs. ham email, healthy vs. unhealthy patients, idiomatic vs. non-idiomatic phrases) it may be that there is a particular outcome that we are interested in identifying instances of, either because of the cost of getting an instance of this class wrong or because instances of this class are rare. For example, in the health domain it is more important to identify patients who are suffering from a disease than to identify patients who do not have a disease\footnote{An error in predicting that a healthy patient is ill is likely to be corrected through follow up tests, whereas an error made in predicting an ill patient is healthy, resulting in the patient being discharged without further tests or treatment, can have disastrous consequences}. In linguistics, it may be that we are interested in identifying instances of a rare linguistic phenomenon. In these contexts the outcome of particular interest is known as the \textit{positive class} and there are a number of evaluation metrics designed to emphasize the ability of a model to correctly identify  instances of this \textit{positive class}. The \textit{F-score} is an example of this type of evaluation metric\footnote{See \cite{kelleher:2015} for an explanation of a range of evaluation metrics (including the F-score).}

The literature on the task of automatic type identification of idioms, more specifically verb and noun idiomatic combinations (VNIC), illustrates the use of these standard model evaluation metrics. Most of the work in this field either uses accuracy (used by \newcite{fazly:2009}) or F-score (used by \newcite{muzny:2013}, \newcite{senaldi:2016}, \newcite{gian17-idiom}) to compare model performance. 
These measures provide an appreciable sense of the reliability of a given model, which is why they are commonly used as evaluation metrics. However, because they focus on evaluation accuracy they are evaluating both the model's ability to score an instance appropriately and the threshold the model uses. We argue that for evaluating a model that will be deployed for pre-filtering in an annotation project this focus is not appropriate. In these contexts the best model is the model that can sequence instances correctly with a clear separation between positive instances (at the start of the list) and negative instances (at the end of the list). In situations where sequencing and class separation are important more important than thresholding, measures such as \textit{gain} and \textit{cumulative gain} become more useful to evaluate model suitability. In the following subsections we illustrate this distinction between classification accuracy and instance ranking using worked examples.

\subsection{Worked example: Classification} 

We introduce our worked example in Table \ref{t:worked}. Let us assume that we have a gold standard test set consisting of 6 phrases (presented in the first column), 3 of which are labeled as idiomatic (shaded green) and 3 of which are labeled as non-idiomatic (shaded red). We have trained two pre-filtering models (M1 and M2) that output a list of candidates ordered according to some sort of measure of confidence in the positive label being correct (be it a probability, or any other type of scaled measure). When using a traditional evaluation metric (accuracy, F-score, etc.) these confidence scores are converted to a class label by applying a threshold to the confidence scores. This cutoff point on our ordered list serves as a delimiter above which we consider all candidates to be labeled as positive (idiomatic), and below which we consider all candidates to be labeled negative (non-idiomatic) by our classifiers. Generally, different models will score instances differentally and will also have different thresholds. Consequently they will return different proportions of positive and negative predictions for a given test set. However, for the purposes of this discussion (and without loss of generality) we will assume that the interactions between the instance scoring and thresholding in each model is such that both models predict that two thirds of the test set are positive examples. Thus $\sim$60\% of our data is labeled as positive by our model (shaded green), while 40\% is labeled as negative (shaded red).

\begin{table}[htbp]
\begin{center}
\begin{tabular}{l||rr||rr}
TestSet & \multicolumn{2}{c||}{\bf S1}  & \multicolumn{2}{c}{\bf S2} \\ 
\hline
\bf ID  & \bf M1 & \bf M2 & \bf M1 & \bf M2 \\
\hline 
\cellcolor{green!65}1 & \cellcolor{green!55}1 & \cellcolor{green!55}1 & \cellcolor{green!55}3 & \cellcolor{green!30}5 \\
\cellcolor{green!65}2 & \cellcolor{green!30}5 & \cellcolor{green!55}2 & \cellcolor{green!30}6 & \cellcolor{green!30}4 \\
\cellcolor{green!65}3 & \cellcolor{green!30}4 & \cellcolor{green!30}6 & \cellcolor{green!30}4 & \cellcolor{green!55}3 \\
\cellcolor{red!70}4 & \cellcolor{green!55}2 & \cellcolor{green!30}5 & \cellcolor{green!30}5 & \cellcolor{green!55}1 \\
\cline{2-5}
\cellcolor{red!70}5 & \cellcolor{red!60}6 & \cellcolor{red!30}3 & \cellcolor{red!30}2 & \cellcolor{red!60}6 \\
\cellcolor{red!70}6 & \cellcolor{red!30}3 & \cellcolor{red!60}4 & \cellcolor{red!30}1 & \cellcolor{red!30}2 \\
\hline
\bf Acc & 0.5 & 0.5 & 0.17 & 0.5 \\
\hline
\end{tabular}
\caption{Two hypothetical prediction scenarios (S1 and S2) that illustrate the classification and evaluation process via thresholding an ordered list. Cells shaded green represent instances classified as positive, while cells shaded red represent instances instances classified as negative. Cells shaded in lighter colors represent false positives and false negatives, while the darker shaded cells represent true positives and true negatives.
\label{t:worked}}
\end{center}
\end{table}

Evaluating this output by comparing to the test set reveals the model's classification accuracy; given that accuracy is a statistic interested in evaluating prediction of both the positive and negative class, when performing the evaluation we divide the correct predictions by the total number of examples in the test set, rather than only the ones in the higher ranked group. 

With this in mind, we present two scenarios: in the first scenario (S1), our first model (M1) and our second model (M2) end up with the same accuracy, meaning their performance is comparable - they both correctly predict the class of 3 out of 6 instances (the correctly labeled instances are shaded darker than the incorrectly labeled instances). Given that they have the same accuracy score, choosing a better model is a non issue. However, in the second scenario (S2), M1 has a lower accuracy than M2, meaning that M1 performs worse than M2. In this scenario, the obvious choice is the model with the higher accuracy - M2.

\subsection{Worked example: Sequencing}

However, when the ordered list returned by a pre-filtering model is used to order the candidates presented to a human annotator and, furthermore, the human annotator will not annotate the entire set of positive predictions (due to budget constraints), then the distribution of correct positive candidates within this ordered list, i.e. the sequence in which they are given to the annotator, becomes very important.

In other words, although two models that return the same number of correct predictions will be judged as identical using classification accuracy metrics, these models may still differ in terms of the distributions of true positive and true negative (i.e. prediction errors) instances: one model may group the true positive instances together near the top of the list and group the true negative instances near the bottom of the list, whereas the other model may intersperse the true negative instances with the true positive instances. For a model used to pre-filter data used in annotation, the model that bunches the true positive instances above the true negative instances is much more useful than the model that intersperses true positives and true negatives because the annotation process will likely be focused on the top portion of the list. This shift in focus is illustrated in Table \ref{t:midpoint}, using the same example as before.

\begin{table}[htbp]
\begin{center}
\begin{tabular}{l||rr||rr}
TestSet & \multicolumn{2}{c||}{\bf S1}  & \multicolumn{2}{c}{\bf S2} \\ 
\hline
\bf ID  & \bf M1 & \bf M2 & \bf M1 & \bf M2 \\
\hline 
\cellcolor{green!65}1 & \cellcolor{green!55}1 & \cellcolor{green!55}1 & \cellcolor{green!55}3 & \cellcolor{red!60}5 \\
\cellcolor{green!65}2 & \cellcolor{red!60}5 & \cellcolor{green!55}2 & \cellcolor{red!60}6 & \cellcolor{red!60}4 \\
\cellcolor{green!65}3 & \cellcolor{red!60}4 & \cellcolor{red!60}6 & \cellcolor{red!60}4 & \cellcolor{green!55}3 \\
\cellcolor{red!70}4 & \cellcolor{green!55}2 & \cellcolor{red!60}5 & \cellcolor{red!60}5 & \cellcolor{green!55}1 \\
\cellcolor{red!70}5 & \cellcolor{red!60}6 & \cellcolor{green!55}3 & \cellcolor{green!55}2 & \cellcolor{red!60}6 \\
\cellcolor{red!70}6 & \cellcolor{green!55}3 & \cellcolor{red!60}4 & \cellcolor{green!55}1 & \cellcolor{green!55}2 \\
\hline
\end{tabular}
\caption{Two hypothetical prediction scenarios (S1 and S2) that illustrate the the difference of considering sequencing of true positive/true negative over classification accuracy. Instances of the positive class are shaded green, whereas the instances of the negative class are shaded red.
\label{t:midpoint}}
\end{center}
\end{table}

As we are not interested in predicting the negative class, we focus our interest only on the instances of the positive class (shaded green). In an ideal scenario, we would employ an annotator to go through the ordered list and pick out the true positive instances so we can add them to our dictionary. If we can go over the whole list, the ordering is still not very important, as we will eventually get to all the positive instances.

However, as is often the case when building a linguistic resource, budgetary constraints mean that we cannot manually review all of the examples that a model returns as positive. Simulating the effect of these constraints on the test set requires a cut off point in the data. Thus, once budget comes into play, our list of candidates is shortened: if we can only afford to annotate $\sim$30\% of our positively labeled data, six candidates drop out of each list, and we drop two thirds of the dataset (shaded light green and light red). This is illustrated in Table \ref{t:thepoint}.

\begin{table}[htbp]
\begin{center}
\begin{tabular}{l||rr||rr}
TestSet & \multicolumn{2}{c||}{\bf S1}  & \multicolumn{2}{c}{\bf S2} \\ 
\hline
\bf ID  & \bf M1 & \bf M2 & \bf M1 & \bf M2 \\
\hline 
\cellcolor{green!65}1 & \cellcolor{green!55}1 & \cellcolor{green!55}1 & \cellcolor{green!55}3 & \cellcolor{red!60}5 \\
\cellcolor{green!65}2 & \cellcolor{red!60}5 & \cellcolor{green!55}2 & \cellcolor{red!60}6 & \cellcolor{red!60}4 \\
\cline{2-5}
\cellcolor{green!65}3 & \cellcolor{red!25}4 & \cellcolor{red!25}6 & \cellcolor{red!25}4 & \cellcolor{green!25}3 \\
\cellcolor{red!70}4 & \cellcolor{green!25}2 & \cellcolor{red!25}5 & \cellcolor{red!25}5 & \cellcolor{green!25}1 \\
\cellcolor{red!70}5 & \cellcolor{red!25}6 & \cellcolor{green!25}3 & \cellcolor{green!25}2 & \cellcolor{red!25}6 \\
\cellcolor{red!70}6 & \cellcolor{green!25}3 & \cellcolor{red!25}4 & \cellcolor{green!25}1 & \cellcolor{green!25}2 \\
\hline
\bf TP & 1 & 2 & 1 & 0 \\
\hline
\end{tabular}
\caption{Two hypothetical prediction scenarios (S1 and S2) that illustrate the advantage provided by considering the distribution of classes in the ordered list. Lighter-shaded cells represent data that cannot be annotated due to budget constraints. The bottom row shows total number of true positive (TP) instances included in the final resource.
\label{t:thepoint}}
\end{center}
\end{table}

At this point, if our goal is to get the largest amount of true positive examples for our resource, the choice between M1 and M2 becomes more complex. In S1, the most profitable choice becomes M2 (even though it has the same accuracy as M1 when judged traditionally), as it will yield 2 entries for our dictionary, whereas M1 would yield only 1 entry for the dictionary.\footnote{Note we are defining the concept of \textit{profit} here as the ratio of resources invested (cost) and positive instances gained (benefit).} In S2, on the other hand, the most profitable choice is M1, as it would provide 1 entry for our dictionary, whereas M2 would provide 0 entries, even though its overall accuracy was almost thrice as high as that of M1.

This is only a provisional example, but it is important to keep in mind that these proportions do scale. If the model remains unchanged, the accuracy and distribution tendencies stay the same, no matter the size of the dataset. Because we know that for a given project budget we can have a human annotator review a limited number of instances, we are interested not so much in which model is the best at identifying positive instances across a full test-set, but rather which model will sequence its output in such a way as to return the largest number of true positive predictions within the top $n$ instances ranked by the model as positive. Thus, accuracy is not an ideal measure in this situation, and relying exclusively on it can have very expensive consequences. In contrast, a measure that is very closely related to the size of the budget is \textit{gain}.

\section{The gain measure}\label{ss:gain}

In Section \ref{s:eval} the concept of a \textit{positive class} was introduced as a category of instances in which we are particularly interested. In contexts where such a positive class exists it is appropriate to use a model evaluation metric that emphasizes the performance of the model on the correctly identified instances belonging to this category. The F-score is an example of an evaluation metric which focuses on evaluating model performance on a positive class. The drawback with using the F-score to evaluate pre-filtering models in an annotation context is that it does not provide any insight into the distribution of true positive and true negative instances within the set of instances predicted to be positive by a model. \textit{Gain}, and the associated concept of \textit{cumulative gain}, also focus on a positive class but have the advantage of taking into account the separation of positive and negative instances:

\begin{quote}
``The basic assumption behind gain is that if we were to rank the instances in a test set in descending order of the prediction scores assigned to them by a well-performing model, we would expect the majority of the positive instances to be toward the top of this ranking. The gain metric attempts to measure to what extent a set of predictions made by a model meets this assumption'' \cite[p. 433]{kelleher:2015} 
\end{quote}

An important difference between traditional evaluation metrics and \textit{gain} is that whereas traditional evaluation metrics return a single score for a model on a test set, using \textit{gain} results in a number of \textit{gain} scores for each model for a test set. In order to calculate the \textit{gain} for a model, we first rank the instances in a test set according to the prediction score returned by the model for each instance. We then divide this ordered list into deciles (groups containing 10\% of the dataset) and calculate the model's \textit{gain} in each decile by dividing the true positive instances in each decile  (based on the known target labels in the dataset) by the total number of positive instances in the test set: 

\begin{equation}
\label{eq:gain}
gain(dec)=\frac{num\: positive\: test\: instances\: in\: decile\: \textit{dec}}{num\: total\: positive\: test\: instances}
\end{equation}

Hence, once we know what the \textit{gain} is for each decile of our ordered list, with our budget dictating how many deciles we can annotate, we can easily identify at which point checking more candidates stops being profitable, or how much data we need to pass to our annotator to obtain the desired number of entries for our dictionary. Comparing these measures across several systems then proves to be a suitable evaluation measure. We can go back to Table \ref{t:thepoint} and apply the \textit{gain} calculation to our hypothetical data. We are assuming that each row/data point is also one decile. Hence, we know that in S1 the first decile (D1) of both M1 and M2 has a \textit{gain} of 0.5. However, the \textit{gain} of M1 in D2 is 0, while the \textit{gain} of M2 is, again, 0.5. Given that we can afford to only annotate 2 deciles, this makes M2 the clear winner.

\subsection{Cumulative gain}

A concept related to \textit{gain} is \textit{cumulative gain}, which can be defined as the proportion of positive instances in a data set that have already been identified up to a given decile. This means that as we go through the deciles, the value of \textit{cumulative gain} rises until the last decile when it reaches its maximum. Plotting the \textit{cumulative gain} values for each decile produces a \textit{cumulative gain chart} (shown in Section \ref{s:casestudy}, Figure \ref{f:gainchart}), which allows us to understand how many of the positive instances in a complete test set we can expect to have identified at each decile of the dataset.  

So in our example from Table \ref{t:thepoint}, in the first decile D1 of S1-M1 the \textit{cumulative gain} is equal to that same decile's \textit{gain} - 0.5. But once we reach D4 (which also has a \textit{gain} of 0.5), its \textit{cumulative gain} will be 1, as by that point we will have encountered all the positive instances in the list. 

This illustrates why we are particularly interested in \textit{cumulative gain}, as it allows us to make the following consideration: for a given budget X we can annotate N deciles of our data. Knowing that model M has the highest \textit{cumulative gain} at that point then tells us that we should use model M. 

Moving on from our hypothetical example, this reasoning can be applied to a number of real natural language processing tasks. We illustrate the application of the \textit{gain} metric in 
the following section where we showcase its usefulness on a real dataset.

\section{Case Study - Idiom Dictionary}\label{s:casestudy}

As mentioned in Section \ref{s:eval}, we examine the usefulness of \textit{gain} on the task of building a dictionary of idioms. In our setting, we have several different predictive models at our disposal that perform type identification of potential idiomatic combinations of verbs and nouns. More specifically, our case study is based on the task of identifying potential verb and noun idiomatic combinations (VNIC). The output of each of our models provides an ordered list of potential candidate phrases for the dictionary, ranked according to a confidence measure. Hence, we do not actually classify the instances; we simply order all the instances presented to the prefiltering model and then pass them in sequence to an expert who checks whether the candidates are idiomatic or not.

We compare the performance of three models that perform VNIC type identification based on fixedness measures \cite{fazly:2009}. The particularities of the three models are irrelevant for this analysis, but it is important to note that the data featured here is actual experimental data stemming from real models.\footnote{For more details on the background of the models, refer to \newcite{gian17-idiom}. The resulting dataset is published on GitHub as a freely available resource. It can be obtained through this link: \url{https://github.com/giancds/vnics_dataset}.}
Their global performance evaluations (precision, recall and F-measure) are presented in Table \ref{t:eval}. 

\begin{table}[htbp]
\begin{center}
\begin{tabular}{l|ccc}
\hline
\bf Model & \bf Pr. & \bf Rec. & \bf F1 \\
\hline 
M1 
& 0.82 & 0.73 & 0.70 \\
M2 
& 0.83 & 0.75 & 0.74 \\
M3 
& 0.83 & 0.78 & 0.77 \\
\hline
\end{tabular}
\caption{Results in terms of precision, recall and F-score, ordered by their F-scores, in the classification task over a balanced test set. The scores were calculated as the weighted average of Precision, Recall and F-Scores over each class (VNICs and non-VNICs).
\label{t:eval}}
\end{center}
\end{table}

The table shows the models' evaluation results over a balanced test set. 
The metrics were calculated as the weighted average of the Precision, Recall and F-Scores on each class (VNICs and non-VNICs). All of the results are statistically significant according to Spearman's ranked correlation test (all $p$ \textless{} 0.05).

\begin{figure}[h!]
	\centering
	\includegraphics[width=0.49\textwidth]{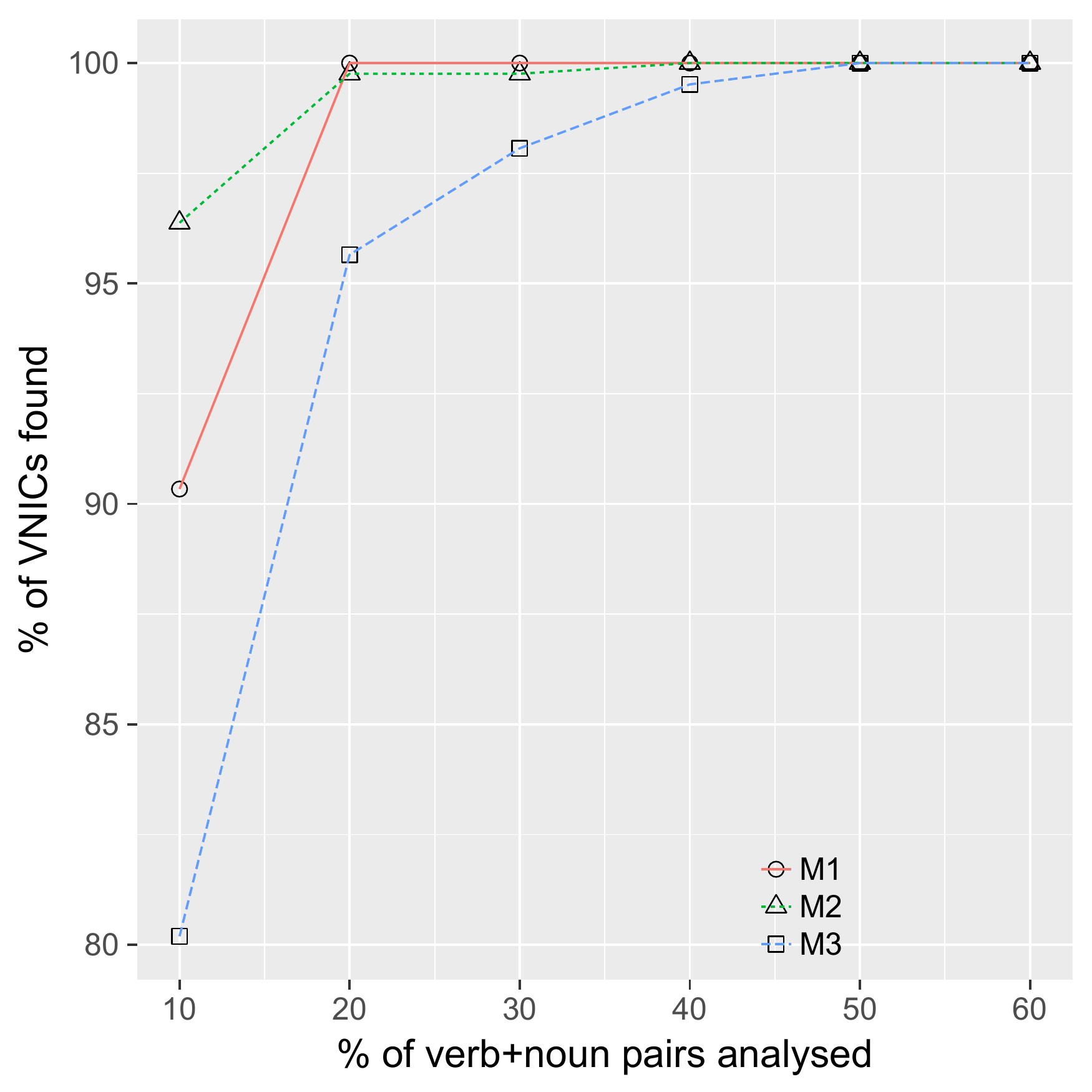}
	\caption{Cumulative gain chart for our three models' performance in the retrieval task.
    \label{f:gainchart}}
\end{figure}

As the table shows, there are variations between the F-scores: the first model (M1) has the lowest F-score, while the third model (M3) has the highest F-score, traditionally making M3 the clear choice. However, our annotation budget is constrained, so the question we ask is: which of the machines will provide the maximum benefit in terms of the number of correctly identified VNICs, given our budget? To answer this, we calculate \textit{gain} and \textit{cumulative gain}, as described in Subsection \ref{ss:gain} Results of the calculations are presented in Figure \ref{f:gainchart}

The graphic relates the percentage of verb and noun pairs that were analyzed with the number of VNICs that were correctly identified by the models, essentially showing the \textit{gain} calculations through each of the model's deciles. There are several perspectives from which we can make interpretations of this graph, depending on the variables of budget (i.e. number of deciles we can afford to annotate, which can be fixed or flexible) and the number of positive instances (i.e. VNICs, which is either 'as many as possible', or an actually targeted number, which is externally dictated).
Consequently, we can find ourselves in one of the following situations:

\paragraph{(Situation 1)} We have a fixed budget and can only afford to annotate a set amount of deciles, in hopes of getting as many VNICs as possible. We are looking for the model that can yield the maximum number of VNICs in the same amount of deciles. The choice of model for this situation then depends on the number of deciles we can annotate, and can even change depending on that number. For example, if we can afford to annotate only one decile, then Figure \ref{f:gainchart} tells us that M2 (painted green) is the preferred choice, as it will yield the most VNICs. However, if we can afford to annotate two deciles (or three, or four), then M1 (painted red) would give us more VNICs than M2, making it the preferred model. Finally, even though it has the highest F-score of the three models, M3 (painted blue) can really only be considered as an option if we can annotate five or more deciles.

\paragraph{(Situation 2)} We have a flexible budget and we can afford to annotate a variable amount of deciles, but we still want to spend as little money as possible, yet find the maximum number of VNICs. In other words, we are looking for the model that can yield the maximum number of VNICs in the earliest decile. In our case study, that would be M1, which reaches the maximum positive instances in the second decile. In comparison, M2 reaches the maximum in the fourth decile, while M3 reaches it only in the fifth decile. This means that, if we were to choose M3, we would have to annotate 50\% of the data to obtain all the candidates for our dictionary, while we would have to annotate only 20\% of the data if we were to choose M1. It is interesting to note that ranking models with this in mind would show that M1\textgreater{}M2\textgreater{}M3, which is inverse to the models' ranking according to their F-scores. Thus, choosing a model based solely on the fact that it has the highest F-score would result in an unnecessary investment of time and money.

\paragraph{(Situation 3)} We have performed some annotation and we have leftover funding - we wonder whether it is worth annotating some more of our data, e.g. another decile.
Figure \ref{f:gainchart} can also help us in deciding whether it is worth spending more money on the human annotation at any given point. Say we have chosen M2 as our model and have already annotated three deciles of the data. We can consult the \textit{cumulative gain chart} to see how many more VNICs we can expect if we decide to invest in annotating an extra decile of the data. 
For M1, we know we would get nothing back after the second decile, as we would already have all the available VNICs, so investing more money there would be pointless. However, if we were working with M3, it is obvious that annotating one further decile is very profitable at every step up until the fourth decile, where the \textit{gain} starts to decline heavily. Such considerations could allow us to avoid unnecessary annotation and save money.

\subsection{Cost model}

There is an annotation cost associated with marking the candidate phrase as idiomatic or not. This cost is assumed to be fixed. The benefit is binary - if the human annotator deems a candidate to be idiomatic, we get a new entry in our dictionary; if it is not deemed idiomatic, there is no benefit, as the annotator was paid for considering the example, but this did not result in expanding the dictionary. 

Similar to \cite{balamurali-12}, we can apply some values to these variables and perform calculations of cost, in order to get a more palpable sense of the benefits of such considerations. To do this, we need to fix an annotation cost associated with annotating a VNIC candidate. Given that annotation costs vary for different tasks in different parts of the world, for the purpose of this illustration we fix the annotation cost at \$0.04 per candidate phrase. This value was chosen as it is the cost of a similar task (annotation of personal nouns) available on Amazon Mechanical Turk\footnote{\url{https://www.mturk.com}}.

Using these values, we can apply the gain measure to perform cost calculations on our test set. We keep in mind that the gain scores scale to different sizes of datasets: if the model had a gain of $x$ on the third decile of the test set, we can extended our annotation to cover the third decile of the project data as ordered by the model, we would be likely to have a similar gain on our larger dataset and ultimate linguistic resource. In other words, just like with gain, we can extrapolate from the cost calculations on our test set to estimate the cost of annotating a larger-scale unlabeled dataset.

We thus turn to our case study, in which our test set is our ordered list of 2,091 verb+noun pairs, of which 414 are VNICs. Using the aforementioned annotation costs, we know that the cost of a single annotator going through 1 decile (10\% of data) would be \$8.3, and annotating the whole list would cost us \$83.64 (100\% of data). Once we have these values ready, our cost calculations depend on what our goals are, as derived from the different situations described in the previous section.

If our goal is the one from Situation 1 -- to get as many VNICs as possible on a fixed budget -- we need to consider the budget and how many VNICs it can give us for each system.
For illustrative purposes, we assume that our annotation budget is fixed at \$16. This means we can afford to annotate not more than two deciles. If we spend that money on M1 we will actually get all 414 VNICs. If we go with M2, we will get slightly less  - 410 VNICs. If we opt for M3, however, we will get back 394 VNICs. This means that the difference between spending \$16 on M1 and M3 is a difference of 20 VNICs, or 5\% fewer positive instances.

If our goal is to maximize the benefit while spending as little money as possible on a flexible budget (as in Situation 2), this means we want a model that will find all the true positives as early as possible. To annotate all the VNICS we would have to spend a minimum of \$16.56 in the ideal scenario where they are all found at the top of the ordered list. However, this is not the case with any of our three models. The costs of annotating all available VNICs in relation to each model are presented in Table \ref{t:costs}. As we can see, comparing the cheapest model M1 with the most expensive model M3 yields a difference of 60\%.

\begin{table}[htbp]
\begin{center}
\begin{tabular}{lc}
\hline
\bf Model & \bf Cost \\
\hline 
M1 
& \$16.73 \\
M2 
& \$33.46 \\
M3 
& \$41.82 \\
\hline
\end{tabular}
\caption{Cost of annotating the maximum amount of true positives for each model on the test set from our case study.
\label{t:costs}}
\end{center}
\end{table}

These values may seem small, but, as already stated in Section \ref{s:eval}, these are calculations performed on a test set, and the proportions do scale to larger samples. As a consequence, in larger projects with vast amounts of data and numerous annotators, saving 60\% of the annotation budget can turn out to be quite a considerable amount.

Finally, if we have already done some annotation, but our goal is to see if it is worth annotating some more data (as in Situation 3), we want to see how many more VNICs we will get if we annotate another decile. Let us assume that we have annotated only the first two deciles. This means we have already spent \$16.6. If we look at M1 and M2, we see that spending another \$8.3 (50\% of the already spent budget) will actually give us 0 new VNICs, so we know that spending more resources on annotation would be fruitless. However, looking at M3's third decile, we see that, if we spent that additional \$8.3, we would in fact get an additional 8 VNICs, which is a much more valuable cost-benefit trade-off than the previous two. 

In addition, this final line of thinking does not exclusively apply to model comparisons. Even in a scenario where we only have a single predictive model at our disposal, rather than comparing different models' performances, we can still perform \textit{gain} calculations to know when to stop annotating the output of just the one model. If M3 is our only available model, we may perform these calculations after each decile, and depending on how frugal we want to be, we could decide that the additional 4 VNICS obtained by annotation the 4th decile are not worth the additional investment of \$8.3, thus saving the funding for other potential annotation tasks.

\section{Conclusion}

In this paper we emphasize budgetary concerns in the task of creating a linguistic resource that includes predictive modeling and subsequent human annotation of the output data. We demonstrate different ways of how using the measure of \textit{gain} can help when faced with a choice between similarly performing models on a limited budget. 

We show that examining the \textit{gain} of a predictive model can help us answer several important budgetary questions. It helps us identify which model will yield the highest profit at the earliest time, or rather, which model will yield the highest number of positive annotations, given our budget. It can also inform us on how much we need to invest to achieve a desired number of entries in our dictionary, as well as tell us at which point annotating more candidates stops being profitable.

We demonstrate all of this on real experimental data, highlighting that sometimes a model that has considerably higher traditional evaluation scores can actually perform much worse in any of the above regards, ending up as more costly than a model with a lower evaluation score. 

For the purpose of this paper we have illustrated our point on the example of building a dictionary of idioms as a linguistic resource, but we should note that the same considerations can be applied to the task of creating any other linguistic resource, or indeed many other NLP tasks, as long as the pipeline integrates a probabilistic model.

For example, in SemEval-2018\footnote{\url{http://alt.qcri.org/semeval2018/}}, one of the shared tasks was hypernym discovery\footnote{\url{https://competitions.codalab.org/competitions/17119}}. The goal was to build a system that would take a word or phrase as its input (e.g. folk rock) and then generate a list of the 10 most likely candidate hypernyms for that word (e.g. rock, genre, music, ...). Given that a word can have more than one correct hypernym, the goal of the task was not only to build a system with high global accuracy and F-scores, but also to build a system that can cleanly separate the positive and negative classes. Because the output of the task is constrained to the top 10 candidates, and those candidates are submitted as potential hypernyms, we would want the system to assign the highest probabilities to the true positives so that, ideally, all or most of the top 10 candidates are indeed hypernyms. Thus, if participants of the task were to build several systems, calculating \textit{gain} will show which system would likely yield the highest evaluation scores, which could help inform their decision on which system to submit for evaluation.

Certainly, we are not claiming that performing \textit{gain} calculations will invariably undermine the results of global F-score evaluations. We only wish to raise awareness of the fact that when it comes to model selection, external constraints (such as budget) can drastically change the perspective on the model's performance, which warrants more specialized consideration. Indeed, the usual evaluation metrics such as accuracy and F-score are a perfectly appropriate tool for an intrinsic comparison of models, but once they are taken out of the lab, the evaluation should be extrinsic, taking into account the whole pipeline of the task the models are applied to. This shift in context warrants a reconsideration of the appropriate evaluation metric, and \textit{gain} might just be the answer.


\section*{Acknowledgements}

This research was supported by the ADAPT Centre for Digital Content Technology which is funded under the SFI Research Centres Programme (Grant 13/RC/2106) and is co-funded under the European Regional Development Fund. Giancarlo D. Salton would like to thank Capes ("Coordena\c{c}\~{a}o de Aperfei\c{c}oamento de Pessoal de N\'{i}vel Superior") for his Science Without Borders scholarship, proc n. 9050-13-2.

\section*{Bibliographic References}

\bibliographystyle{lrec}
\bibliography{xample}


\end{document}